\RequirePackage{xcolor}
\documentclass{article}

\usepackage[final]{corl_2020} % Uncomment for the camera-ready ``final'' version.

\usepackage[utf8]{inputenc} % allow utf-8 input
\usepackage[T1]{fontenc}    % use 8-bit T1 fonts
\usepackage{float}
\usepackage{url}            % simple URL typesetting
\usepackage{booktabs}       % professional-quality tables
\usepackage{amsfonts}       % blackboard math symbols
\usepackage{nicefrac}       % compact symbols for 1/2, etc.
\usepackage{microtype}      % microtypography

\usepackage{algorithm}
\usepackage{algpseudocode}
\usepackage{amsmath}
\usepackage{amssymb,amsfonts,amsthm}
\usepackage{mathtools}  % also loads amsmath
\usepackage{bbm}
\usepackage{pifont}%
\usepackage{graphicx}
\newcommand{\mengye}[1]{} % {\textcolor{orange}{[Mengye: #1]}}
\newcommand{\nicv}[1]{} % {\textcolor{blue}{[Nic: #1]}}
\newcommand{\raquel}[1]{} % {\textcolor{red}{[Raquel: #1]}}

\definecolor{cadmiumgreen}{rgb}{0.0, 0.42, 0.24}
\newcommand{\new}[1]{#1} % {\textcolor{blue}{#1}}

\newcommand{\SUPP}{} % comment out to get rid of supplemental
\newcommand{\supplemental}[1]{
\ifdefined\SUPP
  #1
\else
\fi
}

\DeclareMathOperator*{\argmax}{arg\!\max}

\newcommand{\cmark}{\ding{51}}

\title{Learning to Communicate and Correct Pose Errors}
\newcommand{\pose}{\boldsymbol \xi}
\newcommand{\noisypose}{\tilde{\pose}}
\newcommand{\predpose}{\widehat{\pose}}

\newcommand{\relpose}{\pose}
\newcommand{\noisyrelpose}{\noisypose}
\newcommand{\predrelpose}{\predpose}

\newcommand{\msg}{\mathbf m}
\newcommand{\aggmsg}{\mathbf h}

\newcommand{\edge}[2]{#1#2}  % represents #1->#2

\title{Learning to Communicate and Correct Pose Errors}

\author{
  Nicholas Vadivelu$^{1,2}$\thanks{Work done while at Uber ATG.}, Mengye Ren$^{1,3}$, James Tu$^{1,3}$, 
  \textbf{Jingkang Wang$^{1,3}$ and Raquel Urtasun$^{1,3}$} \\
  \\
  Uber Advanced Technologies Group$^1$, \; University of Waterloo$^2$, \; University of Toronto$^3$ \\
  \texttt{nbvadive@uwaterloo.ca, \{mren3,james.tu,jingkang,urtasun\}@uber.com} \\
}

\begin{document}
\maketitle

%===============================================================================

% !TEX root = ../main.tex
\begin{abstract}
Learned communication makes multi-agent systems more effective by aggregating distributed
information. However, it also exposes individual agents to the threat of erroneous messages they
might receive. In this paper, we study the setting proposed in V2VNet~\cite{v2vnet}, where nearby self-driving
vehicles jointly perform object detection and motion forecasting in a cooperative manner. Despite a
huge performance boost when the agents solve the task together, the gain is quickly diminished in
the presence of pose noise since the communication relies on spatial transformations. Hence, we propose a
novel neural reasoning framework that learns to communicate, to estimate potential errors,
and finally, to reach a consensus about those errors. Experiments confirm that our proposed framework significantly
improves the robustness of multi-agent self-driving perception and motion forecasting systems under realistic and severe localization noise.

\end{abstract}

\keywords{multi-agent, self-driving, perception, prediction}

%===============================================================================

% !TEX root = ../main.tex
\vspace{-0.1in}
\section{Introduction}
Despite the powerful capabilities of deep neural networks in fitting raw, high dimensional data,
they are limited by the computational power and sensory input available to a single agent. Thus,
combining the sensory information and computational power of multiple agents to cooperatively
accomplish a goal can greatly amplify the effectiveness of these systems \cite{v2vnet,
liang_deep_2018, chen_cooper_2019, obst_multi-sensor_2014, rawashdeh_collaborative_2018}. For
example, V2VNet~\cite{v2vnet} has recently shown that by allowing multiple self-driving vehicles
(SDVs) to communicate through a set of learned spatially-aware feature maps, we can obtain
significant gains in detecting obstacles that would have otherwise been occluded or far away from a single-agent
perspective.

The success of V2VNet depends on the precise localization of each participating vehicle, which is
used to warp the feature maps so they can be spatially aligned. Localization noise, however, is
 common in the real world. While V2VNet
exhibits some implicit tolerance, the performance degrades below single-agent performance under
realistic amounts of noise. Due to the safety critical nature of self-driving, it is
paramount to study the robustness against pose noise in a vehicle-to-vehicle communication
system and to design models that can explicitly reason under such noise.

In this paper, we propose end-to-end learnable neural reasoning layers that learn to communicate, to
estimate pose errors, and finally, to reach a consensus about those errors. First, the pose regression module predicts the
relative pose noise between a pair of vehicles. Second, to ensure globally consistent poses, we
propose a consistency module based on a Markov random field with Bayesian reweighting. Lastly, in
the communicated messages aggregation step, we propose using predicted attention weights to attenuate the
outlier messages among vehicles.

Our evaluation   under the same setting as the original V2VNet shows that our  model can
maintain the same level of performance under strong translation and heading localization noise, while V2VNet eventually suffers from such input noise, even if the network is
trained with data augmentation. Our framework also outperforms other competitive pose
synchronization methods.

% !TEX root = ../main.tex

\begin{figure}[t]
  \centering
    \includegraphics[width=0.9\textwidth]{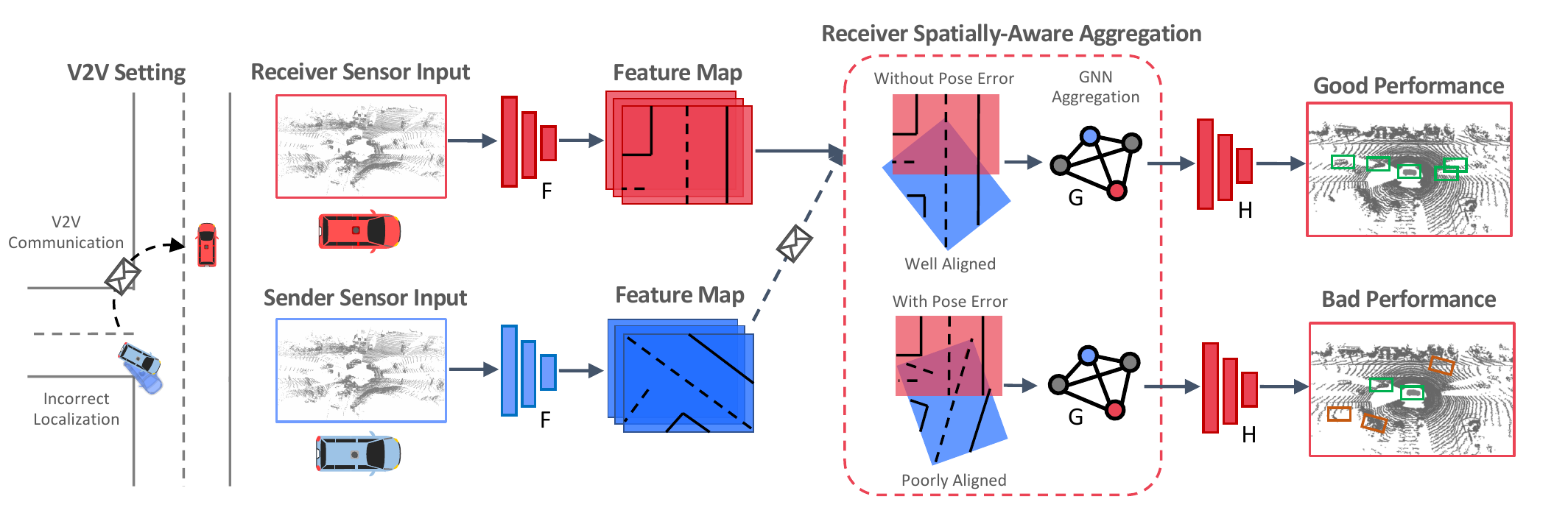}
  \vspace{-0.15in}
  \caption{\textbf{V2V communication setting with pose noise without correction.}
  We demonstrate the case where there is one receiver (\textcolor{red}{red}) and one sender
  (\textcolor{blue}{blue}). Typically, the communication would be two-way, but we illustrate one
  way for clarity. Pose noise causes the features to be misaligned during aggregation, making them
  unusable for detection and motion forecasting tasks.}
  \label{fig:motivation}
  \vspace{-0.15in}
\end{figure}

\section{Related Work}

In this section, we describe the  literature in the area of collaborative self-driving. We
also give an overview on related problem formulations such as transformation synchronization, visual
odometry, and point-cloud registration.

\vspace{-0.1in}
\paragraph{Collaborative self-driving:} Existing literature studies how to leverage multiple
self-driving vehicles (SDVs) to perform vehicle-to-vehicle communication (V2V) to enhance perception,
prediction, and motion planning. The benefits of multiple agents can be exploited by aggregating raw
sensor data~\cite{chen_cooper_2019}, communicating intermediate feature maps~\cite{v2vnet}, or  
combining the outputs of multiple vehicles~\cite{rawashdeh_collaborative_2018,rauch_car2x_2012,
rockl_v2v_2008}. \cite{v2vnet, chen_cooper_2019} show limited robustness to localization error, with
no explicit steps to address it. We follow the setting of V2VNet~\cite{v2vnet} by communicating
intermediate feature maps since it achieves better performance and more efficient communication.

\vspace{-0.1in}
\paragraph{Transformation synchronization:} Transformation synchronization is the process of
extracting absolute poses given relative poses. Methods include spectral solutions
\cite{bernard_solution_2015, arrigoni_spectral_2016,huang_learning_2019, gojcic_learning_2020},
semidefinite relaxations \cite{rosen_certifiably_2020, singer_angular_2009, bernard_solution_2015},
probabilistic approaches \cite{rosen_certifiably_2020, birdal_bayesian_2018}, sparse matrix
decomposition \cite{arrigoni_robust_2018}, and/or learned approaches
\cite{huang_learning_2019,gojcic_learning_2020, purkait_neurora_2019}. While these methods could be
used to refine our pairwise estimates, they are only shown to be robust when there are many
more views to synchronize than in our setting (e.g., 30 views per scene in
\cite{choi_redwood_2016} vs. up to 7 in our setting). Hence, they are susceptible to outliers
which strongly influence the final synchronized poses. Our approach can certainly be used in standard transformation synchronization problems, but more importantly, we propose an
end-to-end system for robust multi-agent perception and motion forecasting.

\vspace{-0.1in}
\paragraph{Visual odometry:} Visual odometry is the process of determining the pose of an agent
given images from the agent's view. In our setting, when correcting pose error, we extract the
relative poses given pairs of views. \citet{yousif_overview_2015} provide a survey on several
visual odometry methods, including feature-based \cite{talukder_real-time_2003,
dornhege_visual_2006} and stereo-based \cite{matthies_error_1990, kess_flow_2009}. More
recently, approaches based on  RCNN \cite{mohanty_deepvo_2016, wang_deepvo_2017} learn this task
end-to-end. These approaches are optimized for images, LiDAR, or other raw sensory inputs, whereas
in our setting, we aim to align intermediate feature maps.

\vspace{-0.1in}
\paragraph{Point cloud registration:} Point cloud registration is the task of finding a (typically
rigid) transformation to align two point clouds. \cite{yang_polynomial_2019, fitzgibbon_robust_2002}
propose robust methods for registration, while \cite{lu_deepicp_2019,yew_3dfeat-net_2018} propose
deep learning based approaches. \cite{pomerleau_review_2015} provides a full review of traditional
point cloud registration methods. These methods are not suited in our setting due to the high communication overhead
required to transmit LiDAR point clouds to neighboring self-driving vehicles.

\vspace{-0.1in}
\paragraph{Multi-agent deep learning:} Outside of self-driving, there is broad literature on
multi-agent deep learning systems. \cite{omidshafiei_deep_2017, balachandar_collaboration_2019}
communicate actions and state to other agents, while \cite{sukhbaatar_learning_2016} use a
controller network for communication. Our setting is more similar to the former, where each vehicle
communicates an intermediate representation of its view to nearby vehicles. 
\cite{marvin} uses a learned graph neural network for communication and cooperative routing. 
However, many of these
methods are typically studied in toy settings, whereas we evaluate our model on a realistic
self-driving dataset.
% !TEX root = ../main.tex
\begin{figure}[t]
  \centering
    \includegraphics[width=0.9\textwidth,trim={1.2cm, 6cm, 2cm, 2cm},clip]{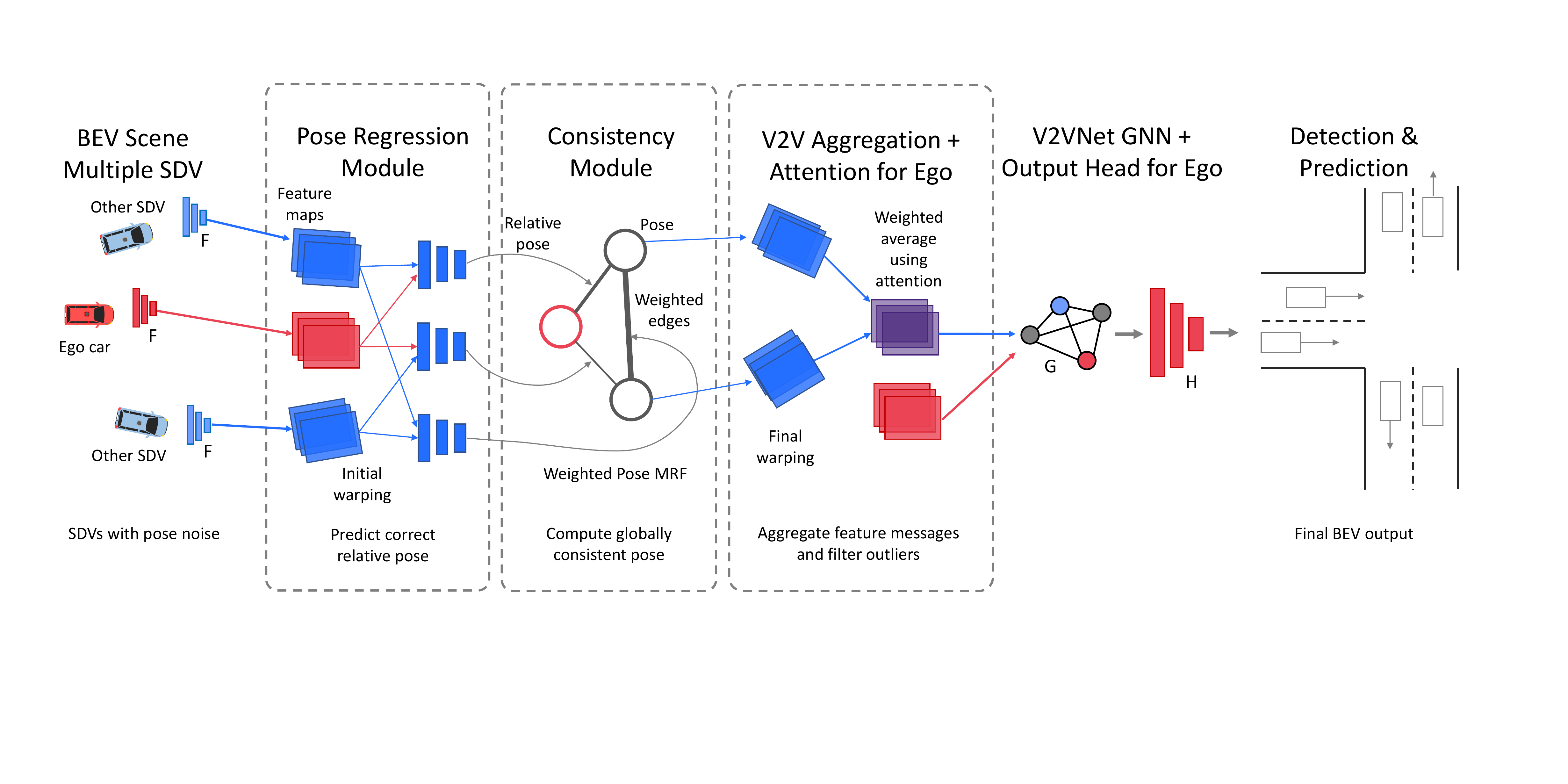}
    \vspace{-0.1in}
  \caption{\textbf{Our proposed method for robust V2V communication under pose error.}
  The network's feature maps are communicated in the style of V2VNet~\cite{v2vnet}, but before the
  final warping step, we propose end-to-end learnable modules. First, the \textit{pose regression
  module} and the \textit{consistency module} to fix pose errors. Lastly, before aggregation, the
  \textit{attention module} predicts a soft binary attention weight used in a weighted average of
  messages to filter out remaining noisy messages. In contrast, V2VNet performs a uniform average
  instead of a weighted average during the GNN step.}
  \label{fig:mainfig}
  \vspace{-0.1in}
\end{figure}

\section{Learning to Communicate and Correct Pose Errors}
\vspace{-0.1in}

Pose noise has been shown to severely detriment existing collaborative multi-agent self-driving
systems. In this section, we describe our novel approach to correct pose errors in such settings.
In the following, we first review V2VNet \cite{v2vnet}, the collaborative self-driving framework that we base our
models on. We then propose a pose error correction network composed of i) a pose regression module
to predict pairwise relative poses, ii) a consistency module to reach global consensus, and iii) an
attention aggregation module to filter out outlier messages. These modules are learned end-to-end
jointly to improve object detection and motion forecasting.

\vspace{-0.1in}
\subsection{Background on V2VNet}
Our pose correction approach is based on V2VNet  \cite{v2vnet}, a state-of-the-art collaborative multi-vehicle
self-driving network which has been shown to provide significant improvements in both object detection and motion
forecasting over single vehicle systems. We call the combined detection and forecasting task
perception and prediction (PnP). We first review the background of V2VNet---an overview diagram
is illustrated in Figure~\ref{fig:motivation}.

\vspace{-0.1in}
\paragraph{Input parameterization and message computation:}
Given multiple LiDAR sweeps, V2VNet  voxelizes the point cloud into   15 cm$^3$  voxels,
and concatenates them along the height dimension to form a birds-eye view input representation. It then processes this representation using a 2D CNN, denoted $F$, to produce a spatial feature
map of shape $c \times l \times w$ (channels, length, width).  To facilitate cooperation, each
self-driving vehicle (SDV) compresses and broadcasts these spatial feature maps to nearby SDVs. We thus call these spatial feature maps
\textit{messages} and denote \new{the message from vehicle i as} $\msg_{i}$.
%\new{\sout{ for agent i}}.

\vspace{-0.1in}
\paragraph{Message \new{passing and }aggregation:}
\new{Vehicle $i$ collects all incoming messages and aggregates them via a graph neural network (GNN) $G$~
\cite{li_gated_2017}. The set of vehicles which communicate with vehicle i is denoted $adj(i)$.}
When vehicle $i$ receives message $\msg_j$ \new{from vehicle $j \in adj(i)$}, it warps $\msg_j$ from the
perspective of vehicle $j$ to its own. Vehicle i uses its own pose $\pose_i$ and the other vehicle's pose
$\pose_j$ to compute the relative pose $\relpose_{\edge{j}{i}}$. % 
The message from vehicle j ($\msg_j$) is transformed via $\relpose_{\edge{j}{i}}$ to produce the warped message $\msg_{\edge{j}{i}}$, which is aligned to the perspective of vehicle i. 
Let the aggregated message for agent $i$ be $\aggmsg_i :=
G\left(\{\msg_{\edge{j}{i}}\}_{j \in adj(i)} \right)$ 
We refer to \cite{v2vnet} for details on the aggregation algorithm.

\vspace{-0.1in}
\paragraph{Output parameterization and header:}
Finally, vehicle $i$  uses a CNN $H$ to process aggregated messages to predict the final outputs which
consist of object detections represented with their  3D position, width, height, and orientation, as well as
prediction outputs representing the locations of objects at future time steps. 

\vspace{-0.1in}
\paragraph{Learning objective:} V2VNet is trained using the PnP loss
$\mathcal L_{PnP} (\mathbf {y_i}, \mathbf {\hat y_i})$, which is a combination of a cross-entropy
loss on the vehicle classifications, smooth $\ell_1$ on the bounding boxes, and smooth $\ell_1$ on
the predicted motion forecasting waypoints.

\vspace{-0.1in}
\paragraph{Pose notation:} \new{
Since processing is done in birds-eye view, poses are in $SE(2)$. We represent
each pose as a vector $\pose \in \mathbb R^3$, consisting of two translation components and one rotation angle.
We denote composing two transformations via $\pose_1 \circ \pose_2$, which is equivalent to
multiplying their corresponding homogeneous transformation matrices. We denote $\mathbf \pose^{-1}$ as
the inverse pose, equivalent to inverting the corresponding transformation matrix}

\subsection{Robust V2V communication against pose noise}

V2VNet has been shown to be vulnerable to pose noise because misaligned incoming messages will
result in unusable features for the network. Under realistic noise, V2VNet's performance can be
worse than single vehicle PnP. In this section we introduce details of our approach to improve
robustness against pose noise. An illustration is shown in Figure~\ref{fig:mainfig}.

In our setting, each SDV $i$ has a noisy estimate of its own pose denoted $\noisypose_i$, and
receives the noisy poses of neighboring self-driving vehicles as part of the messages. These
noisy poses are used to compute the noisy relative transformation from SDV $j$ to $i$
denoted $\noisyrelpose_{\edge{j}{i}}$.

\paragraph{Pose regression module:} \label{regression}
Since all the vehicles perceive different views of the same scene, we use a CNN to learn the
discrepancy between what a vehicle sees and the orientation of the warped incoming messages. The
network for the i-th agent takes $(\msg_i \mathbin\Vert \msg_{\edge{j}{i}})$ as input and outputs a
correction $\mathbf{\widehat c}_{\edge{j}{i}}$ such that $\mathbf{\widehat c}_{\edge{j}{i}} \circ
\noisyrelpose_{\edge{j}{i}} = \predrelpose_{\edge{j}{i}}$. $\mathbin\Vert$ denotes concatenation
along the features dimension, and $\mathbf{\widehat c}_{\edge{j}{i}} \circ
\noisyrelpose_{\edge{j}{i}}$ represents applying the transformation $\mathbf{\widehat
c}_{\edge{j}{i}}$ to the noisy relative transformation $\noisyrelpose_{\edge{j}{i}}$, to produce a predicted true relative transformation $\predrelpose_{\edge{j}{i}}$. Note that
since we make an independent prediction for each directed edge, $\predrelpose_{\edge{j}{i}} \neq
\predrelpose_{\edge{i}{j}}^{-1}$. In our setting, concatenating the features at the input was shown
empirically to be more effective than using an architecture with two input branches that are
concatenated downstream (which is done in \cite{luo_efficient_2016, agrawal_learning_2015}).

\paragraph{Consistency module:}
We now refine the relative pose estimates from the regression module by finding a set of globally
consistent absolute poses among all our SDVs. By allowing the SDVs to reach a global consensus about
eachothers absolute pose, we can further mitigate pose error.

We formulate our consistency as a Markov random field (MRF), where each vehicle pose is a node and we condition on the
predicted relative poses. Since the predicted relative pose error will have many outliers, 
the distribution of our true absolute poses conditioned on these will have a heavy tail. We thus
assume each pose $\pose_i$ follows a multivariate student $t$-distribution with mean $\pose_i \in
\mathbb R^3$ and scale $\Sigma_i \in R^{3 \times3}$ conditioned on the relative poses.
We do not use any unary potentials.
Our pairwise potentials consist of three components: the likelihoods, weights, and weight priors: 
\begin{align}
\psi(i,j) = &
  \underbrace{p(\predrelpose_{\edge{j}{i}} \circ \pose_j)^{w_{\edge{j}{i}}}
             p(\predrelpose_{\edge{j}{i}}^{-1} \circ \pose_i)^{w_{\edge{j}{i}}}}_{
             \text{Weighted Likelihood given } \predrelpose_{\edge{j}{i}}}
  \underbrace{p(\predrelpose_{\edge{i}{j}} \circ \pose_i)^{w_{\edge{i}{j}}}
             p(\predrelpose_{\edge{i}{j}}^{-1} \circ \pose_j)^{w_{\edge{i}{j}}}}_{
             \text{Weighted Likelihood given } \predrelpose_{\edge{i}{j}}}
  \underbrace{p(w_{\edge{j}{i}}) p(w_{\edge{i}{j}})}_\text{Weight Priors}.
\end{align}

The likelihood terms $p(\predrelpose_{\edge{j}{i}} \circ \pose_j)$ and
$p(\predrelpose_{\edge{i}{j}}^{-1} \circ \pose_j)$, both $t$-distributed centered at $\pose_i$,
encourage the result of the relative transformation ($\predrelpose_{\edge{j}{i}}$ or $\predrelpose_{\edge{i}{j}}^{-1}$) from a source vehicle position ($\pose_j$) to stay close to the target
vehicle's position ($\pose_i$). Both directions are included due to symmetry of the rigid transformations. However, not all
pairwise transformations provide the same amount of information, and since our regression module
tends to produce heavy tailed errors, we would like to reweight the edge potentials to downweight
erroneous pose regression outputs. Concretely, we use a weight $w_{\edge{j}{i}} \in \mathbb R$  for each
term in the pairwise potential: $p(\predrelpose_{\edge{j}{i}}
\circ \pose_j)^{w_{\edge{j}{i}}}$, so that low weighted terms will influence the estimates less. We use a prior
distribution for each $w_{\edge{j}{i}}$, where the mean of the distribution is $o_{\edge{j}{i}} \in \mathbb R$---the fraction of spatial
overlap between two messages. Intuitively, we would like to trust the pose prediction more if the
two messages have more spatial overlap. Following \cite{wang_robust_2017}, we use a Gamma prior:
$p(w_{\edge{j}{i}}) =
\Gamma(w_{\edge{j}{i}} \; | \; o_{\edge{j}{i}}, k)$, where $k$ is the shape parameter.

\begin{algorithm}
\caption{Consistency module inference}
    \begin{algorithmic}[1]
        \State $\pose_i \gets \noisypose_i$ \quad $i = 1...n$
        \State $w_{\edge{j}{i}} \gets 1$ \quad $(i, j) \in \mathcal E$
        \For {$k = 1...$\text{num\_iters}}
            \State $\pose_i, \Sigma_i \gets \argmax_{\pose_i, \Sigma_i} \mathcal
                   \prod_{j\in adj(i)} p(\predrelpose_{\edge{j}{i}} \circ \pose_j)^{w_{\edge{j}{i}}} \;
                                       p(\predrelpose_{\edge{i}{j}}^{-1} \circ \pose_j)^{w_{\edge{i}{j}}}$
                   \quad $i = 1...n$ \label{argmax_t}
            \State $w_{\edge{j}{i}} \gets \argmax_{w_{\edge{j}{i}}} p(w_{\edge{j}{i}} \; | \; \pose_i, \Sigma_i$) \quad $(i, j) \in \mathcal E$ \label{argmax_w}
        \EndFor \\
    \Return $\pose_i$ \quad $i=1...n$
    \end{algorithmic}
    \label{alg:consistency}
\end{algorithm}

To perform inference on our MRF, we would like to estimate the values of our absolute poses
$\pose_i$, the scale parameters $\Sigma_i$, and the weights $w_{\edge{j}{i}}$ that maximize the
product of all our pairwise potentials. We achieve this via Iterated Conditional Modes
\cite{besag_statistical_1986}, described in Algorithm~\ref{alg:consistency}.
The maximization step on Line~\ref{argmax_t} happens simultaneously for all nodes via weighted
expectation-maximization (EM) for the $t$ distribution \cite{liu_ml_1999}. We provide the EM
algorithm in the Supplementary Material. The maximization step on Line~\ref{argmax_w} can be
computed using the following closed form~\cite{wang_robust_2017}:
\begin{align}
  \argmax_{w_{\edge{j}{i}}} p(w_{\edge{j}{i}} \; | \; \pose_i, \Sigma_i) = \frac{o_{\edge{j}{i}} k}
                      {k - \log p(\predrelpose_{\edge{j}{i}} \circ \pose_j) -
                           \log p(\predrelpose_{\edge{j}{i}}^{-1} \circ \pose_i)}.
\end{align}
We then use these estimated poses to update the relative transformations needed to warp the messages.

\paragraph{Attention aggregation module:}
After we predict and refine the relative transformations, there may still be errors present in some
messages that hinder our SDVs' performance. In V2VNet, warped incoming messages are averaged when
being processed by the GNN $G$. This means each message will make an equal contribution towards the
final predictions. Instead, we want to focus on clean messages and ignore noisy ones. Thus, we
propose a simple yet effective attention mechanism to assign a weight to each warped message before
they are averaged, to suppress the remaining noisy messages. We use a CNN $A$ to predict an
unnormalized weight $s_{\edge{j}{i}} \in \mathbb R$. Specifically, $\text{sigmoid}(A(\msg_i
\mathbin\Vert \msg_{\edge{j}{i}})) = s_{\edge{j}{i}}$. We compute the normalized weight
$a_{\edge{j}{i}} \in \mathbb R$ as follows:
\begin{align}
a_{\edge{j}{i}} = \frac{s_{\edge{j}{i}}}{\alpha + \sum_{k \in adj(i)} s_{\edge{k}{i}}}.
\end{align}
The learned parameter $\alpha \in \mathbb R$ allows the model to ignore all incoming messages if
needed.  Without $\alpha$, if all the incoming messages are noisy, thus all the $s_{\edge{j}{i}}$
are small, the resulting weights would be large after the normalization. Then, we can compute our
aggregated message: 
\begin{align}
\aggmsg_i = G\left(\{ a_{\edge{j}{i}} \msg_{\edge{j}{i}}
\}_{j \in adj(i)}\right).
\end{align}
The aggregated message is then used by the network $H$ to predict bounding boxes for object detection and waypoints at future timesteps for motion forecasting.

\subsection{Learning}
\paragraph{Supervising attention:} We first train V2VNet and the attention network. We treat
identifying noisy examples as a supervised binary classification task, where clean examples get a
high value and noisy examples get a low value. For the data and labels, we generate and apply strong
pose noise to some vehicles and weak pose noise to others within one scene.
Concretely, we generate the noise via $\mathbf n_i \sim \mathcal D_w$ or $\mathbf n_i \sim \mathcal
D_s$, where $D_w$ is a distribution of weak pose noises, and $D_s$ of strong noises. Like the poses, the noises have
two translational components and a rotational component, thus $\mathbf n_i \in \mathbb R^3$. A fixed
proportion $p$ of our agents receive noise from the strong distribution while the rest from the
weak one. When considering a message, it is considered clean when both agents have noise from the
weak distribution, and considered noisy when either vehicle has noise from the strong distribution.
This labeling is summarized in the following function:
\begin{align}
\text{label}(j, i) =
    \begin{cases}
        \gamma & \mathbf n_j \sim \mathcal D_w \text{ and } \mathbf n_i \sim \mathcal D_w,\\
        1 - \gamma & \mathbf n_j \sim \mathcal D_s \text{ or } \mathbf n_i \sim \mathcal D_s.\\
    \end{cases}
\end{align}
This function produces smooth labels to temper the attention module's predictions so the
attention weights are not just 0 or 1. We define the loss for our joint training task as follows:
\begin{align}
\mathcal L_{joint}(\mathbf y_i, \mathbf{\hat y}_i, \{s_{\edge{j}{i}}\}_{j \in adj(i)}) =
\lambda_{PnP} L_{PnP}(\mathbf y_i, \mathbf{\hat y}_i) +
\frac{\lambda_{attn}}{|adj(i)|}\sum_{j \in adj(i)} \mathcal L_{CE}(\text{label}(j, i), s_{\edge{j}{i}}),
\end{align}
where $\mathcal L_{CE}$ is binary cross entropy loss. This additional supervision was paramount to training the attention mechanism---training with $L_{PnP}$ alone produced a significantly less effective model.

\paragraph{Pose regression:} Then, we freeze V2VNet and the attention and train only the regression
module using $\mathcal L_{c}$. In this stage, all the SDVs get noise from the strong noise
distribution $\mathcal D_s$. We train this network using a loss which is a sum of losses over each
coordinate:
\begin{align}
\mathcal L_{c}(\relpose_{\edge{j}{i}}, \predrelpose_{\edge{j}{i}}) =
\sum_{k=1}^3\lambda_{k}
\mathcal L_{sl1}(\relpose_{\edge{j}{i}}, \predrelpose_{\edge{j}{i}})_k,
\end{align}
with $\lambda =
[\lambda_{pos},
\lambda_{pos}, \lambda_{rot}]$, and $\mathcal L_{sl1}$  the smooth $\ell_1$ loss. This regression formulation was empirically more effective than discretizing the predictions and formulating the problem as classification.

Finally, we fine-tune the entire network end-to-end with the combined loss: $\mathcal L =
\mathcal L_{c} + \mathcal L_{task}$, which is possible because our MRF inference algorithm is
differentiable via backpropogation.
% !TEX root = ../main.tex
\begin{figure}[t]
  \centering
  \includegraphics[width=0.9\textwidth,trim={0 0 0 0.78cm},clip]{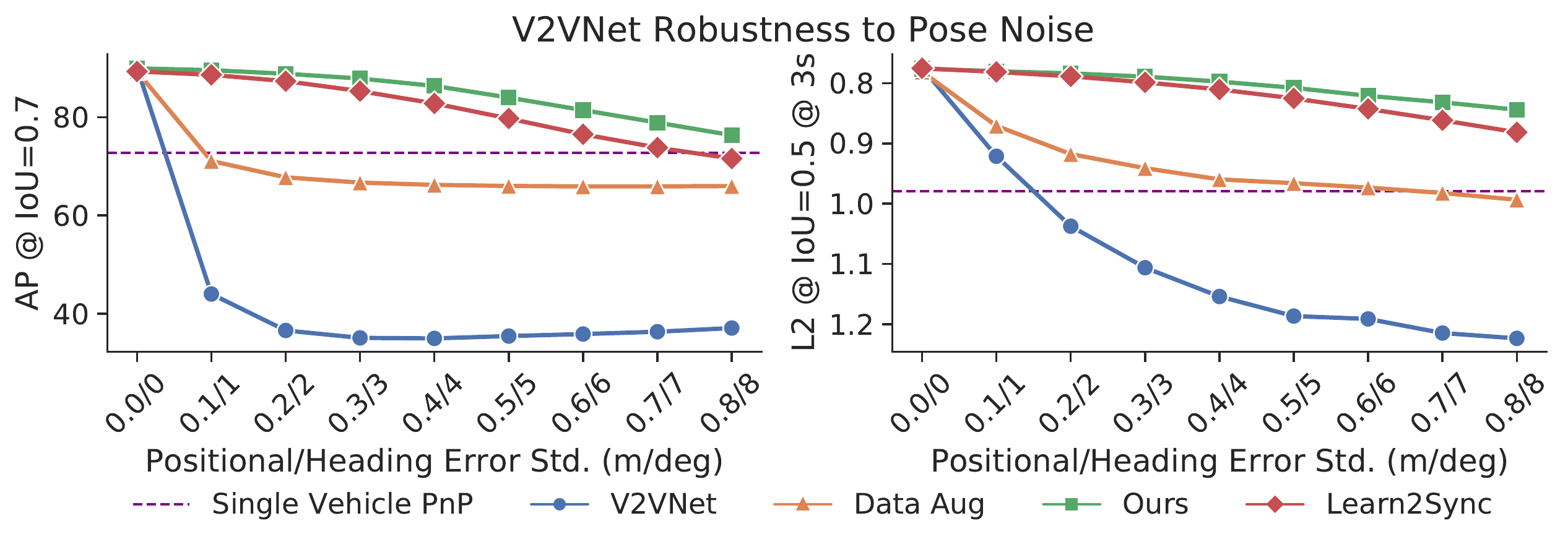}
  \vspace{-0.15in}
  \caption{\textbf{Detection and motion forecasting performance of models across various noise
  levels.} Single Vehicle PnP denotes V2VNet with no SDV peers. Other methods are all trained with
  0.4m/$4.0^\circ$ standard deviation positional/heading noise.}
  \label{fig:pnp_eval}
\end{figure}

\section{Experiments}
\vspace{-0.1in}
We evaluate our method on detection, prediction, and pose correction in various noise settings,
including noise not seen during training. The specific architectures and hyperparameters are provided in the supplementary material.

\subsection{Experimental setup}

\paragraph{Dataset:}
Our model is trained on the V2V-Sim dataset \cite{v2vnet}, which is generated from a high-fidelity
LiDAR simulator \cite{manivasagam_lidarsim_2020}. The simulator uses real-world snippets to first
reconstruct 3D scenes with static and dynamic objects, then simulates LiDAR point clouds from the
viewpoints of multiple self-driving vehicles. Each scene contains up to 7 SDVs. There are
46,796/4,404 frames for the train/test split, where each frame contains 5 LiDAR sweeps. We refer
readers to \cite{v2vnet} for more details.

\vspace{-0.1in}
\paragraph{Evaluation metrics:} Following 
\cite{v2vnet}, detection is measured using Average Precision (AP) at an Intersection
over Union (IoU) of 0.7, motion forecasting (prediction) performance is measured using $\ell_2$
displacement error of the object's center location at a future time step (e.g., 3s in the future) on
true positives. A true positive is a detection where the IoU threshold is 0.5 and the confidence threshold is set such that the recall is 0.9 (we pick the highest recall if 0.9 cannot be reached).
Pose correction performance is evaluated using mean absolute error (MAE) and root mean
squared error (RMSE). All reported metrics are for vehicles in coordinate view range of $x\in
[-100m, 100m]$, $y\in[-40m, 40m]$ around the SDV, which includes objects that are completely occluded (0 LiDAR
points hit), making the task more difficult and realistic. The communicating vehicles themselves are excluded in
evaluation (as PnP of these would be trivial for the co-operative network).

\vspace{-0.1in}
\paragraph{Noise simulation:} Throughout training and evaluation, the noise is sampled and applied
independently to the pose of each SDV. This can be applied as a post-processing step on the data, or
can be simulated directly within LiDARSim \cite{manivasagam_lidarsim_2020}. During training, the
positional noise is drawn from a Gaussian with $\mu = 0$, $\sigma = 0.4$ for $\mathcal D_s$ and
$\sigma = 0.01$ for $\mathcal D_w$; the rotational noise is drawn from a von Mises distribution with
$\mu = 0$, $\sigma = 4^{\circ}$ for $\mathcal D_s$ and $\sigma = 0.1^{\circ}$
for $\mathcal D_w$. During evaluation, the parameters of these distributions are varied as described
for each experiment. We show experiments on both noise similar or greater than the noise levels seen during training. Self-driving cars utilize geometric registration algorithms that localize the vehicle online with respect to a 3D HD map. These methods are very precise, with 99\% of the errors being much smaller than 0.2m, which informed the evaluation ranges chosen.

\vspace{-0.1in}
\paragraph{Competitive method:} We compare our method to a competitive transformation synchronization
method Learn2Sync~\cite{huang_learning_2019}, which considers the pairs of depth maps to iteratively
reweight pairwise registrations when finding globally consistent poses. To process pairs of messages
instead of depth maps, a larger version of the Learn2Sync architecture is used (see supplementary material). 
During evaluation, Learn2Sync is used in place of our consistency module.
Our pretrained pose correction module produces the initial pairwise registrations for Learn2Sync. 

\vspace{-0.1in}
\paragraph{Data augmentation baseline:} For a simple baseline to our method, V2VNet is trained
with noisy poses as a form of input data augmentation, which asks the network to implicitly handle
pose noise instead of explicitly correcting the noise. We refer to this network as \textit{Data
Aug}.

\subsection{Experimental results}

\paragraph{PnP performance:} As shown in Figure~\ref{fig:pnp_eval}, V2VNet is quite vulnerable to pose
noise, especially heading noise. When trained with \emph{data augmentation}, the model becomes
significantly more robust, however, this is at the cost of worse performance in less noisy conditions. The
original model trusts incoming messages too much, whereas the data augmented model trusts them too
little and discards too much information.
Using the correction provides significant benefits: there is little drop in performance when faced with
the  noise seen in the training set (0.4 m, $4.0^\circ$ std.). The model generalizes well to noise
stronger than seen in the training set. Our consistency method shows considerable improvement
over Learn2Sync, which is expected in this case as synchronization algorithms are commonly designed
and evaluated on far larger graphs. Having so few transformations to synchronize renders these
methods vulnerable to  outliers.

\begin{table}[t]
  \centering

  \resizebox{0.9\textwidth}{!}{
  \begin{tabular}{lrrrrrrrr}
    \toprule
    & \multicolumn{4}{c}{Position Error (m)} &
      \multicolumn{4}{c}{Rotation Error (deg)} \\
    & \multicolumn{2}{c}{0.4 m, $4^{\circ}$ std.} &
      \multicolumn{2}{c}{0.8 m, $8^{\circ}$ std.} &
      \multicolumn{2}{c}{0.4 m, $4^{\circ}$ std.} &
      \multicolumn{2}{c}{0.8 m, $8^{\circ}$ std.} \\
                          & RMSE &    MAE &  RMSE &   MAE &  RMSE &   MAE &   RMSE &   MAE \\
    \midrule
    No Correction         & 2.556 & 1.554 & 5.723 & 4.571 & 5.079 & 3.115 & 11.483	& 9.157 \\
    Learn2Sync            & 0.394 & 0.191 & 0.516 & 0.281 & 1.664 & 0.766 &  2.750 & 1.420 \\
    Pairwise              & 0.587 & 0.211 & 0.707 & 0.307 & 2.083 & 0.743 & 3.112 & 1.209 \\
    Gaussian No Reweighting& 0.391 & 0.185 & 0.492 & 0.265 & 1.602 & 0.726 & 2.623 & 1.303 \\
    Gaussian w/Reweighting& 0.283 & 0.153 & 0.377 & 0.218 & 1.386 & 0.634 & 2.379 & 1.153 \\
    Ours \\
    $\to$ Regression Only & 0.644 & 0.245 & 0.825 & 0.377 & 2.186 & 0.803 & 3.275 & 1.326 \\
    $\to$ No Reweighting  & 0.249 & 0.128 & 0.340 & 0.187 & 1.160 & 0.465 & 1.853 & 0.819 \\
    $\to$ Ours            & \bf{0.197} & \bf{0.119} & \bf{0.284} & \bf{0.172} & \bf{0.983} & \bf{0.416} & \bf{1.623} & \bf{0.721}  \\
    \bottomrule
  \end{tabular}}
    \caption{\textbf{Error of the predicted corrections $\mathbf{\hat{c}}_{\edge{j}{i}}$ (as defined in
  \ref{regression}).} No correction corresponds to predicting $\mathbf{\hat{c}}_{\edge{j}{i}} =
  \mathbf{0}$, and is listed to provide context for the metrics. Pairwise refers to averaging the
  relative poses of reverse edges (i.e $(i,j)$ and $(j, i)$). Gaussian refers to our consistency
  formulation with multivariate normal nodes instead of $t$-distributed nodes. No Reweighting
  refers to our consistency formulation without the robust Bayesian reweighting.
  }
  \label{table:correction}
  \vspace{-0.1in}
  
\end{table}

\begin{table}[t]
  \centering
  \resizebox{0.9\textwidth}{!}{
  \begin{tabular}{cccrrrrrr}
    \toprule
    \multicolumn{3}{c}{Modules} & \multicolumn{3}{c}{AP @ IoU = 0.7 $\uparrow$} & \multicolumn{3}{c}{$\ell_2$ @ IoU=0.5 @ 3s $\downarrow$} \\
    Regression & Consistency & Attention & 0.0 / 0 & 0.4 / 4 & 0.8 / 8 & 0.0 / 0 & 0.4 / 4 & 0.8 / 8  \\
    \midrule
            &        &        & 90.070      & 34.960      & 37.065      & 0.774      & 1.154      & 1.223 \\
     \cmark &        &        & 87.777      & 77.227      & 60.312      & 0.793      & 0.830      & 0.901 \\
     \cmark & \cmark &        & 88.906      & 82.241      & 60.978      & 0.787      & 0.813      & 0.884 \\
            &        & \cmark & \bf{90.375} & 67.726      & 67.591      & \bf{0.768} & 0.957      & 0.973 \\
     \cmark &        & \cmark & 89.094      & 84.023      & 75.976      & 0.784      & 0.812      & 0.853 \\
     \cmark & \cmark & \cmark & 89.931      & \bf{86.357} & \bf{76.331} & 0.776      & \bf{0.797} & \bf{0.844} \\
    \bottomrule
  \end{tabular}}
    \caption{\textbf{Ablation of each component of our correction system.} 0.4 / 4 indicates 0.4 m and
  $4^\circ$ standard deviation of noise for position and rotation, respectively. The model with
  none of the modules is V2VNet. Each component provides improvement, with the combination
  of the three producing the best model at high and very high noise.}
  \label{table:ablation}
\vspace{-0.2in}
\end{table}

\vspace{-0.1in}
\paragraph{Pose correction performance:} Table \ref{table:correction} shows that our consistency
module further enhances the correction performance. Note that while the RMSE decreases significantly
with other methods, the MAE only decreases marginally. This implies that, while the outliers are
corrected, the average correction is not improved significantly. Also, this means outliers
``poison'' the good predictions, resulting in relative pose estimates that are mediocre. Improving
the average case is more important than dealing with outliers as our model with
attention can ignore outliers and focus on well-aligned messages.

\vspace{-0.1in}
\paragraph{Ablation studies:} Table \ref{table:ablation} shows that all the
components provide significant benefits. Interestingly, using the attention module provides
improvement over V2VNet even when no noise is present.

\begin{figure}[t]
%   \vspace{-0.15in}
  \centering
    \includegraphics[width=0.9\textwidth,trim={0 0 0 0.77cm},clip]{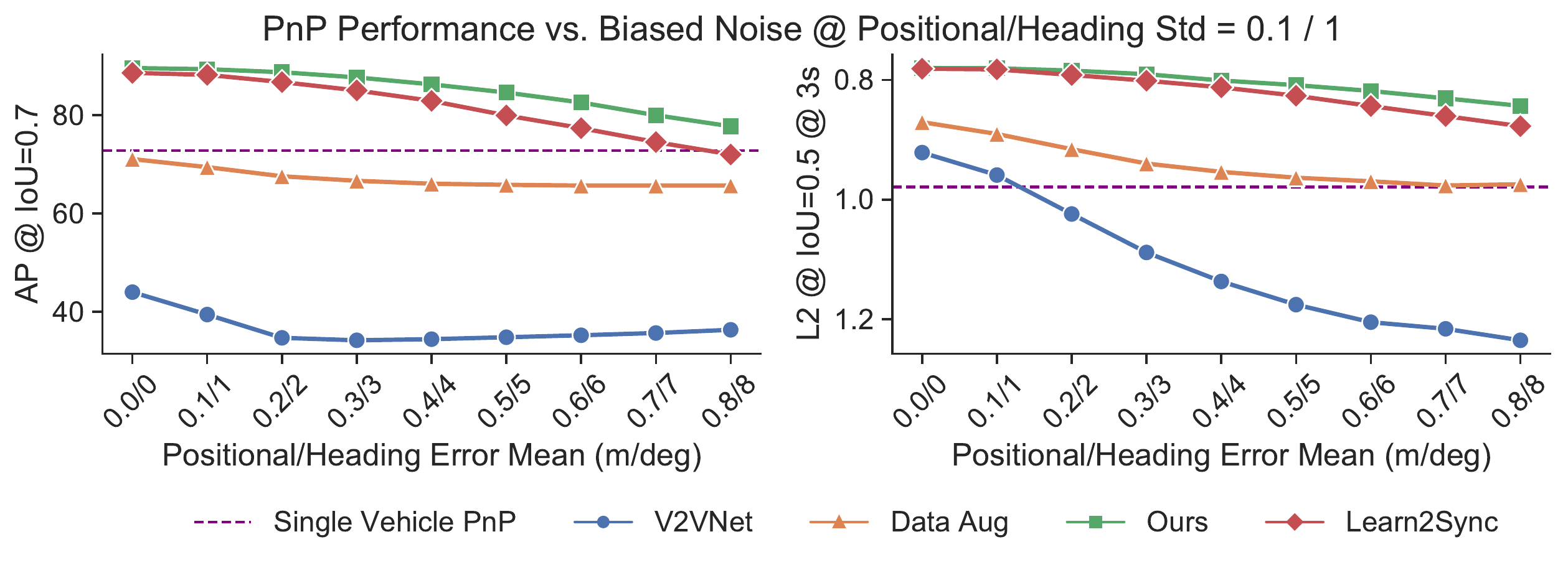}
    \vspace{-0.15in}
  \caption{\textbf{Evaluation of the models against \emph{biased Gaussian noise}}, where the bias is
  varied on the x-axis and the standard deviation is fixed (0.1 m and 1.0$^\circ$). The performance
  stays well above single vehicle PnP, despite the noise being stronger and of an unseen type during
  training.}
  \label{fig:biased_noise}
  \vspace{-0.1in}
\end{figure}
\begin{figure}[t]
  \centering
    \includegraphics[width=0.9\textwidth,trim={0 0 0 0.78cm},clip]{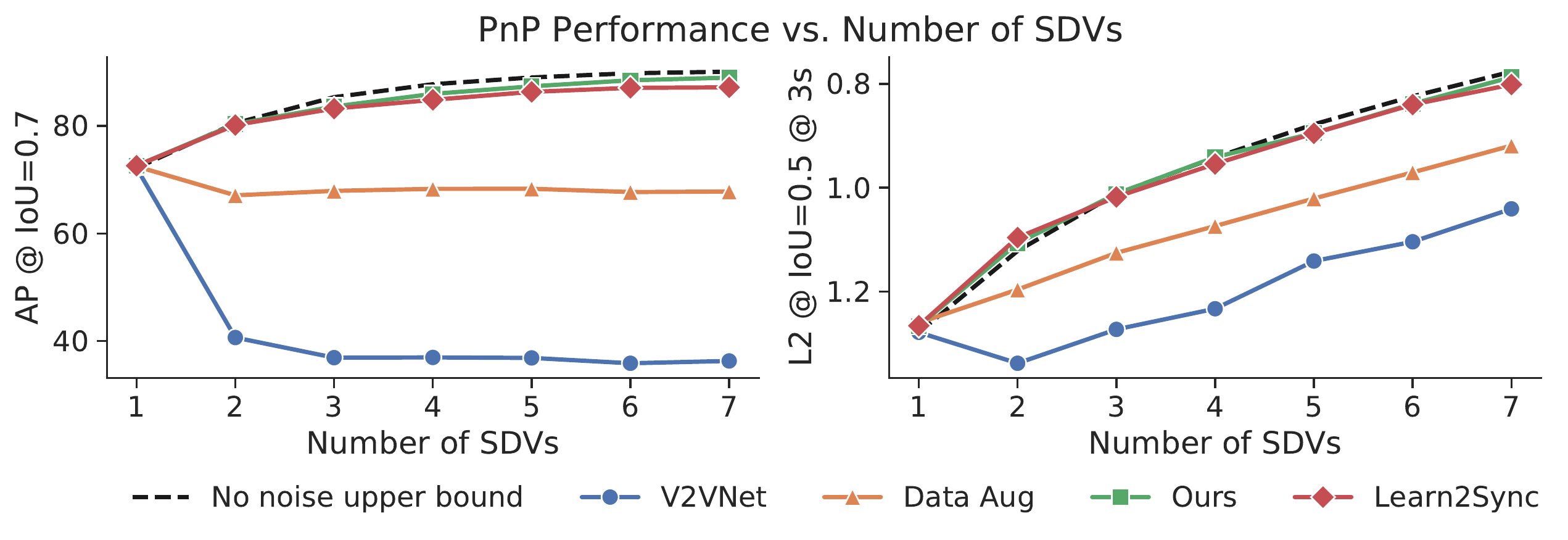}
  \vspace{-0.15in}
  \caption{\textbf{Performance for different numbers of SDV peers.}
  The no noise upper bound is V2VNet evaluated with no noise. The positional/ rotational noise has standard
  deviation 0.2 m / 2$^\circ$.}
  \label{fig:num_veh}
\end{figure}

\vspace{-0.1in}
\paragraph{Biased noise:} There will always be a domain gap between the noise seen during training
and the noise an agent may experience in the real world. In our setting, the pose regression is
trained on unbiased Gaussian noise, however, in the real world, a vehicle may experience
systematic, biased error. Figure \ref{fig:biased_noise} evaluates the generalization ability of
our method on noise that is biased and stronger than what the model may face in reality. The
performance of the model stays well above single vehicle PnP. Furthermore, outliers become more
prevalent in this setting, which affects the performance of consistency methods not designed to
deal with outliers in small graphs.

\vspace{-0.1in}
\paragraph{Number of SDVs:} Strong performance independent of the number of nearby SDVs is
important for safe operation of an SDV. Figure \ref{fig:num_veh} shows that V2VNet's performance
drops as soon as we introduce another SDV due to the pose noise affecting messages, even after
Data Augmentation. This is not the case with our correction: increasing the number of SDVs
improves performance, almost matching the original model evaluated with no noise. The
consistency also maintains reliable performance even with few nearby SDVs.
% !TEX root = ../main.tex
\section{Conclusion}
\vspace{-0.1in}
Collaborative self-driving cars will bring the safety of self-driving to the next level. In this
paper, we propose a collaborative self-driving framework that is made robust to pose errors in
vehicle-to-vehicle communication. Unlike traditional pose synchronization methods, our model is
end-to-end learned to improve detection and motion forecasting. We demonstrate the effectiveness of
our method under various levels of pose noise using V2V simulation. 
In the future, we can extend our work to exploit the temporal consistency of the pose error in incoming messages to improve performance and efficiently reuse computation. We also aim to expand our neural reasoning framework to correct more general types of communication noises to make collaborative self-driving more robust.

%===============================================================================

% The maximum paper length is 8 pages excluding references and acknowledgements, and 10 pages including references and acknowledgements

\clearpage
% The acknowledgments are automatically included only in the final version of the paper.

% \acknowledgments{If a paper is accepted, the final camera-ready version will (and probably should) include acknowledgments. All acknowledgments go at the end of the paper, including thanks to reviewers who gave useful comments, to colleagues who contributed to the ideas, and to funding agencies and corporate sponsors that provided financial support.}
\acknowledgments{We would like to thank Andrei B$\hat{\text{a}}$rsan and Pranav Subramani for insightful discussions. We would also like to thank all the reviewers for their helpful comments.}

%===============================================================================

\bibliography{biblio}  % .bib

\supplemental{
  \newpage
  % !TEX root = ../main.tex

\appendix

\section{EM for weighted $t$-distribution}
Recall in Algorithm~\ref{alg:consistency} on line~\ref{argmax_t} from the
main manuscript we maximize the following quantity for each $i$:

\begin{align}
\argmax_{\pose_i, \Sigma_i} \mathcal \prod_{j\in adj(i)} p(\predrelpose_{\edge{j}{i}} \circ \pose_j)^{w_{\edge{j}{i}}} \;
                   p(\predrelpose_{\edge{i}{j}}^{-1} \circ \pose_j)^{w_{\edge{i}{j}}}.
\end{align}

This is equivalent to finding the weighted maximum likelihood estimate (MLE)
of $\pose_i, \Sigma_i$ given observations $\{\predrelpose_{\edge{j}{i}} \circ
\pose_j\}_{j\in adj(i)} \cup \{\predrelpose_{\edge{i}{j}}^{-1} \circ
\pose_j\}_{j\in adj(i)}$. Recall that $\pose_i, \Sigma_i$ are the location
and scale of the $t$ distribution with $\nu$ degrees of freedom. We modify the EM
algorithm given in \cite{liu_ml_1999} to compute the weighted MLE.

The student $t$ distribution can be defined as follows:
\begin{align}
   p(\predrelpose_{\edge{j}{i}} \circ \pose_j \; | \; \pose_i, \Sigma_i, \nu) =
   \int_0^\infty \mathcal N \left(\predrelpose_{\edge{j}{i}} \circ \pose_j \; | \; \pose_i, (1/\eta_{\edge{j}{i}})\Sigma_i \right)
   \text{Gamma}\left(\eta_{\edge{j}{i}} \; | \; 1, (2/\nu) \right)
   \mathrm d\eta_{\edge{j}{i}},
\end{align}
where 1 is the mean of the Gamma, $2/\nu$ is the shape parameter k, and
$\mathcal N$ denotes the multivariate normal distribution. We provide the
full expressions for the $t$ and Gamma distributions in section
\ref{distributions}. For the expectation step, we compute the expectation of
our latent parameter $\eta_{\edge{j}{i}}$. For the maximization step, we
compute $\pose_i, \Sigma_i$ given $\eta_{\edge{j}{i}}$. We use
$\boldsymbol{\delta}_{\edge{j}{i}}$ to denote the difference between
observation $\edge{j}{i}$ and the current estimate of $\pose_i$ for
convenience. The full algorithm is described in Algorithm~\ref{alg:em_t}.

\begin{algorithm}
\caption{Weighted MLE of $\pose_i, \Sigma_i$ .}
    \begin{algorithmic}[1]
        \State $\pose_i \gets $ \Call{CoordinatewiseMedian}{$\{\predrelpose_{\edge{j}{i}} \circ
\pose_j\}_{j\in adj(i)} \cup \{\predrelpose_{\edge{i}{j}}^{-1} \circ
\pose_j\}_{j\in adj(i)}$}
        \State $\Sigma_i \gets I_{3\times 3}$
        \ForAll {$j \in adj(i)$}
            \State $\boldsymbol{\delta}_{\edge{j}{i}} \gets \pose_i - \left(\predrelpose_{\edge{j}{i}} \circ \pose_j\right)$
            \State $\boldsymbol{\delta}_{\edge{i}{j}} \gets \pose_i - \left(\predrelpose_{\edge{i}{j}}^{-1} \circ \pose_j\right)$
        \EndFor
        \For {$1...$\text{num\_iters}}\\
            \Comment{Expectation Step}
            \ForAll {$j \in adj(i)$}
                \State $\eta_{\edge{j}{i}} \gets \frac{\nu + 3}{\nu + \boldsymbol{\delta}_{\edge{j}{i}}^\top \Sigma_i^{-1}\boldsymbol{\delta}_{\edge{j}{i}}}$
                \State $\eta_{\edge{i}{j}} \gets \frac{\nu + 3}{v + \boldsymbol{\delta}_{\edge{i}{j}}^\top \Sigma_i^{-1}\boldsymbol{\delta}_{\edge{i}{j}}}$
            \EndFor \\
            \Comment{Maximization Step}
            \State $\pose_i \gets \frac{ \sum_{j \in adj(i)}
                                  \eta_{\edge{j}{i}}{w_{\edge{j}{i}}} \left(\predrelpose_{\edge{j}{i}} \circ \pose_j\right) +
                                  \eta_{\edge{i}{j}}{w_{\edge{i}{j}}} \left(\predrelpose_{\edge{i}{j}}^{-1} \circ \pose_j\right)}
                     {\sum_{j \in adj(i)} \eta_{\edge{j}{i}}{w_{\edge{j}{i}}} + \eta_{\edge{i}{j}}{w_{\edge{i}{j}}} } $
            \ForAll {$j \in adj(i)$}
                \State $\boldsymbol{\delta}_{\edge{j}{i}} \gets \pose_i - \left(\predrelpose_{\edge{j}{i}} \circ \pose_j\right)$
                \State $\boldsymbol{\delta}_{\edge{i}{j}} \gets \pose_i - \left(\predrelpose_{\edge{i}{j}}^{-1} \circ \pose_j\right)$
            \EndFor
            \State $\Sigma_i \gets \frac{1}{2|adj(i)|} \sum_{j \in adj(i)} \eta_{\edge{j}{i}} \boldsymbol{\delta}_{\edge{j}{i}}\boldsymbol{\delta}_{\edge{j}{i}}^\top +
                                                                            \eta_{\edge{i}{j}} \boldsymbol{\delta}_{\edge{i}{j}}\boldsymbol{\delta}_{\edge{i}{j}}^\top $ \label{est_cov}
        \EndFor \\
    \Return $\pose_i, \Sigma_i$
    \end{algorithmic}
    \label{alg:em_t}
\end{algorithm}

When there are only two vehicles communicating, we a simple average instead
of EM to estimate $\pose_i$. Notice on line \ref{est_cov} we do not use the
weights $w_{\edge{j}{i}}$, as the small size of our graph often leads to a
singular $\Sigma_i$ when using these weights. 15 iterations is sufficient for
convergence and 2 degrees of freedom worked well.

\section{Additional Experiments}

\begin{figure}[t]
  \centering
    \includegraphics[width=0.9\textwidth,trim={0 0 0 0cm},clip]{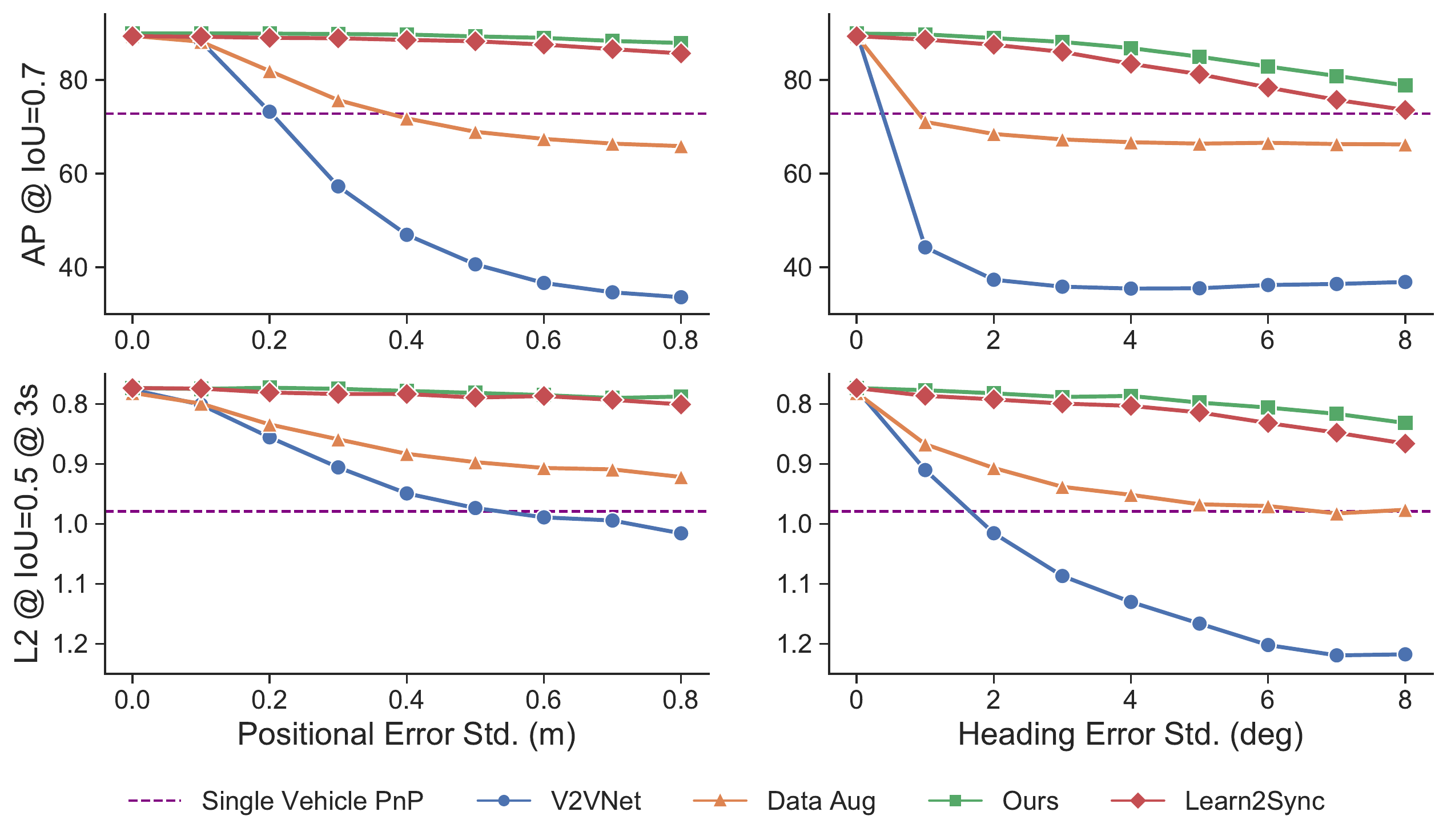}
    \vspace{-0.15in}
  \caption{\textbf{Evaluation of the models against seperated heading and positional noise}.}
  \label{fig:rot_pos_sep}
\end{figure}

We analyze the effects of positional and heading noise seperately in
Figure~\ref{fig:rot_pos_sep}. Heading noise is far more detrimental than positional
noise, as objects far from the vehicle can be displaced significantly even
with slight heading error.

\section{Qualitative Examples}

\begin{figure}[t]
  \centering
  \includegraphics[width=\textwidth,trim={0 25cm 0 0.5cm},clip]{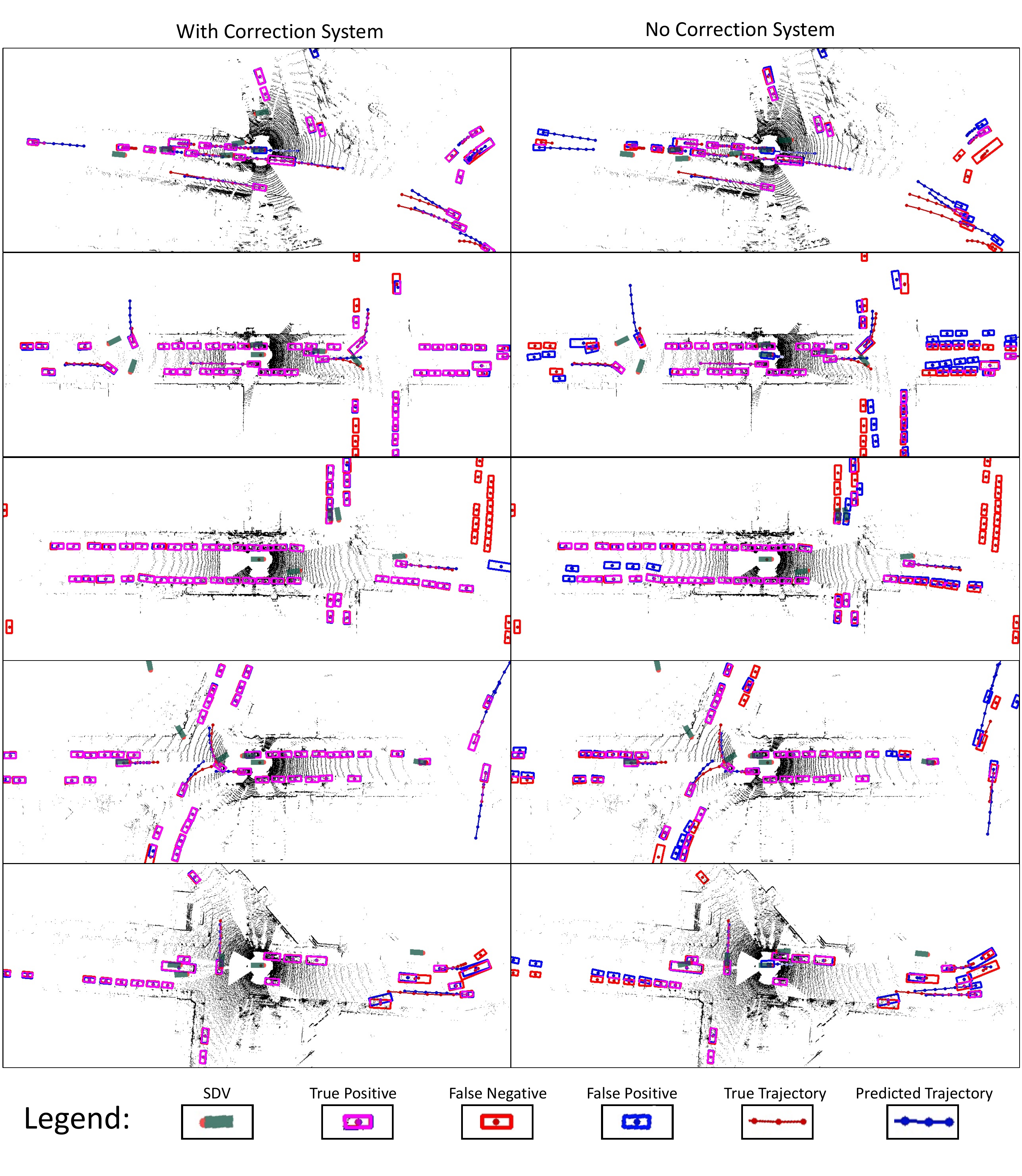}
  \vspace{-0.2in}
  \caption{\textbf{Examples of Perception and Prediction Outputs.} All the
  agents were subject to random pose noise with 0.4 m and 4$^\circ$ standard
  deviation.}
  \label{fig:detection_eg}
\end{figure}

Figure~\ref{fig:detection_eg} shows PnP outputs from five scenes in the
validation set when the agents are subject to pose noise. As shown, the misaligned
messages causes many detections to be innacurate, particularly detections
farther away from the ego vehicle. We also see that forecasting predictions
are skewed without the correction module.

\section{Implementation Details}

In this section, we provide the implementation details for the training procedure and architectures used.

\subsection{Training Hyperparameters}
V2VNet and the attention network are trained using the Adam
optimizer \cite{kingma_adam_2014} with a one-cycle learning rate \cite{smith_super-convergence_2018}
for 6 epochs starting from the pre-trained LiDAR backbone with a peak one-cycle learning rate of
0.0004. Then, V2VNet and the attention network are frozen and only the regression module is trained
for 12 epochs with a peak one-cycle learning rate of 0.002. For the loss, we use $\lambda_{pos} =
2/3$ and $\lambda_{rot} = 1/3$. Finally, the entire network is fine tuned with the combined loss
$\mathcal L$ for 3 epochs with a peak learning rate of 0.0001. For the consistency module, using a
$t$-distribution with 2 degrees of freedom, $k=120$ for the prior worked well, 15 iterations of EM
for the $t$-distribution, 15 steps of ICM, and 10 reweighting steps worked well. The attention
module is trained with $\gamma = 0.9$, $p=0.5$, $\lambda_{PnP} = 0.9$, and $\lambda_{attn} = 0.1$
without significant tuning. We make slight modifications to V2VNet detailed in the supplementary
materials. These modifications resulted in virtually no change in PnP performance.

\subsection{Changes to V2VNet}

Due to GPU memory limitations, we use a slightly altered V2VNet with near
identical performance to the architecture from \cite{v2vnet}. V2VNet
originally performed 3 rounds of message passing between vehicles per
inference; we reduce this to 2. Our correction system only operates during
the first round of propagation. The second round uses the corrected
localization and attention weights from the first round. When receiving
messages, V2VNet uses a convolutional neural network to process each incoming
message before aggregating and passing to the ConvGRU in the GNN. We remove
this processing step and aggregate the messages directly before passing them
to the ConvGRU. Finally, V2VNet uniformly samples between 1 and 7 SDVs per
training example. We sample exactly 4 SDVs per training example when training
V2VNet and the attention, for more consistent GPU memory utilization. We
sample up to 7 SDVs per scene when training only the regression module (as
some training examples have fewer than 7 vehicles).

\subsection{Architecture for our Method}

The dimensions of a message are $(c,l,w) = (80, 128, 320)$. Therefore, the
dimensions of the input to the regression and attention modules are $(160,
128, 320)$. Architectures are described in terms of PyTorch~\cite{paszke_pytorch_2019} modules.
All convolutional layers have a padding
and stride of $(1, 1)$ unless otherwise specified. We annotate each layer with the
output activation shape.

We describe our attention architecture below.
\begin{verbatim}
Sequential(
    Conv2d(160, 160, kernel_size=(3, 3)) -> (160, 128, 320),
    LeakyReLU(negative_slope=0.01) -> (160, 128, 320),
    MaxPool2d(kernel_size=2, stride=2, padding=0) -> (160, 64, 160),
    Conv2d(160, 160, kernel_size=(3, 3)) -> (160, 64, 160),
    LeakyReLU(negative_slope=0.01) -> (160, 64, 160),
    MaxPool2d(kernel_size=2, stride=2, padding=0) -> (160, 32, 80),
    AdaptiveMaxPool2d(output_size=1) -> (160, 1, 1),
    Flatten() -> (160,)
    Linear(in_features=160, out_features=1, bias=True) -> (1,)
)
\end{verbatim}

The use of \texttt{AdaptiveMaxPool2d} is
important: it allows our computed attention weights to be invariant to the
amount of spatial overlap between two messages.

We describe the architecture of our regression module below.
\begin{verbatim}
Sequential(
    Conv2d(160, 160, kernel_size=(3, 3)) -> (160, 128, 320)
    LeakyReLU(negative_slope=0.01) -> (160, 128, 320)
    MaxPool2d(kernel_size=2, stride=2, padding=0) -> (160, 64, 160)
    Conv2d(160, 160, kernel_size=(3, 3)) -> (160, 64, 160)
    LeakyReLU(negative_slope=0.01) -> (160, 64, 160)
    MaxPool2d(kernel_size=2, stride=2, padding=0) -> (160, 32, 80)
    Conv2d(160, 160, kernel_size=(3, 3)) -> (160, 32, 80)
    LeakyReLU(negative_slope=0.01) -> (160, 32, 80)
    MaxPool2d(kernel_size=2, stride=2) -> (160, 16, 40)
    Conv2d(160, 160, kernel_size=(3, 3), stride=(2, 2)) -> (160, 8, 20)
    LeakyReLU(negative_slope=0.01) -> (160, 8, 20)
    MaxPool2d(kernel_size=2, stride=2) -> (160, 4, 10)
    Conv2d(160, 160, kernel_size=(3, 3), stride=(2, 2)) -> (160, 2, 5)
    LeakyReLU(negative_slope=0.01) -> (160, 2, 5)
    MaxPool2d(kernel_size=2, stride=2, padding=0) -> (160, 1, 2)
    AdaptiveMaxPool2d(output_size=1) -> (160, 1, 1)
    Flatten() -> (160,)
    Linear(in_features=160, out_features=160, bias=True) -> (160,)
    LeakyReLU(negative_slope=0.01) -> (160,)
    Linear(in_features=160, out_features=160, bias=True) -> (160,)
    LeakyReLU(negative_slope=0.01) -> (160,)
    Linear(in_features=160, out_features=3, bias=True) -> (3,)
)
\end{verbatim}

\subsection{Architecture for Learn2Sync}

We train Learn2Sync for 10 epochs using the Adam optimizer and a one-cycle
learning with a maximum learning rate of 0.01. We searched for the optimal
learning rate from the set $\{0.1, 0.01, 0.001, 0.0001\}$. Learn2Sync
originally used a modified AlexNet architecture \cite{krizhevsky_imagenet_2012}.
We simply increased the size as detailed
below. The rest of the hyperparameters were kept from \cite{huang_learning_2019}.

\begin{verbatim}
Sequential(
    Conv2d(160, 160, kernel_size=(7, 7), stride=(4, 4)) -> (160, 31, 79)
    ReLU() -> (160, 31, 79)
    LocalResponseNorm(5, alpha=0.0001, beta=0.75, k=2) -> (160, 31, 79)
    MaxPool2d(kernel_size=3, stride=2, padding=0) -> (160, 15, 39)
    Conv2d(160, 256, kernel_size=(5, 5), padding=(2, 2)) -> (256, 15, 39)
    ReLU() -> (256, 15, 39)
    LocalResponseNorm(5, alpha=0.0001, beta=0.75, k=2) -> (256, 15, 39)
    MaxPool2d(kernel_size=3, stride=2, padding=0) -> (256, 7, 19)
    Conv2d(256, 256, kernel_size=(3, 3)) -> (256, 7, 19)
    ReLU() -> (256, 7, 19)
    Conv2d(256, 256, kernel_size=(3, 3)) -> (256, 7, 19)
    ReLU() -> (256, 7, 19)
    AdaptiveMaxPool2d(output_size=(2, 2)) -> (256, 2, 2)
    Flatten() -> (1024,)
    Dropout(p=0.5, inplace=False) -> (1024,)
    Linear(in_features=1024, out_features=1024, bias=True) -> (1024,)
    ReLU() -> (1024,)
    Dropout(p=0.5, inplace=False) -> (1024,)
    Linear(in_features=1024, out_features=1024, bias=True) -> (1024,)
    ReLU() -> (1024,)
    Linear(in_features=1024, out_features=1, bias=True) -> (1,)
)
\end{verbatim}

\section{Distributions} \label{distributions}
We define the $t$-distribution with location $\pose_i \in \mathbb R^3$, scale $\Sigma_i \in \mathbb R^{3\times 3}$, and degrees of freedom $\nu \in \mathbb R$ below. Note that $\pose_i$ is the mean when $\nu > 1$, and $\Sigma_i$ is proportional to the covariance when $\nu > 2$.
\begin{align}
   p(\mathbf{x} \; | \; \pose_i, \Sigma_i, \nu) =
   \frac{\Gamma \left(\frac{\nu + 3}{2}\right)}{\Gamma \left(\frac{\nu}{2}\right)}
   \left(\frac{\left| \Sigma_i^{-1} \right|}{\pi \nu}\right)^\frac{1}{2}
   \left(1 + \frac{(\mathbf x - \pose_i)^T\Sigma_i^{-1}(\mathbf x - \pose_i)}{\nu}\right)^{-\frac{\nu +3}{2}}.
\end{align}

The Gamma distribution with mean $\mu \in \mathbb R$ and shape $k \in \mathbb R$ is defined below:
\begin{align}
   \text{Gamma}(x \; | \; \mu, k) =
   \frac{1}{\Gamma(k) \left(\frac{\mu}{k}\right)^k} x^{k-1} e^{-\frac{kx}{\mu}}.
\end{align}
}

\end{document}